\documentclass{article}
\usepackage{tcolorbox}
\tcbuselibrary{listings,theorems}

\usepackage[numbers]{natbib}

\usepackage[preprint]{neurips_2023}

\usepackage{wrapfig}
\usepackage{graphicx}
\usepackage{booktabs}

\usepackage{bm}
\usepackage{caption}
\newtcbtheorem[]{exmp}{Example}
{colback=purple1!5,colframe=purple1!80,fonttitle=\bfseries, left=.02in, right=.02in,bottom=.02in, top=.02in,fontupper=\small}{exmp}
\usepackage{multicol}
\usepackage{multirow}
\usepackage{tcolorbox}
\usepackage[utf8]{inputenc} % allow utf-8 input
\usepackage[T1]{fontenc}    % use 8-bit T1 fonts
\usepackage{hyperref}
\hypersetup{colorlinks,allcolors=black}
\usepackage{url}            % simple URL typesetting
\usepackage{booktabs}       % professional-quality tables
\usepackage{amsfonts}       % blackboard math symbols
\usepackage{nicefrac}       % compact symbols for 1/2, etc.
\usepackage{microtype}      % microtypography
\usepackage{xcolor}         % colors
\usepackage{colortbl}

\title{DialogGen: Multi-modal Interactive Dialogue System for Multi-turn Text-to-Image Generation}

\author{Minbin Huang$^{1*}$ \quad Yanxin Long$^{2}$\thanks{Equal contribution} \quad Xinchi Deng$^{3}$ \quad Ruihang Chu$^{1}$ \quad Jiangfeng Xiong$^{2}$  \\  \textbf{Xiaodan Liang}$^{3}$ \quad \textbf{Hong Cheng}$^{1}$ \quad \textbf{Qinglin Lu}$^{2}$\thanks{Corresponding Author} \quad \textbf{Wei Liu}$^{2}$\\
{\normalsize $^{1}$Chinese University of Hong Kong \quad $^{2}$Tencent Hunyuan } \\
{\normalsize $^{3}$Shenzhen Campus of Sun Yat-sen University }\\
{\tt\small huangminbin@link.cuhk.edu.hk, \{yestinlong, jefxiong, qinglinlu\}@tencent.com,  	}\\
{\tt\small  dengxch5@mail2.sysu.edu.cn, rhchu@cse.cuhk.edu.hk, xdliang328@gmail.com,  } \\
{\tt\small  hcheng@se.cuhk.edu.hk, wl2223@columbia.edu  } \\
}

\definecolor{purple1}{HTML}{8d3a94}
\definecolor{purple}{HTML}{B36A6F}
\begin{document}
\maketitle

\begin{center}
    \label{fig:DialogGen}
    \captionsetup{type=figure}
    \vspace{-5mm}
    \includegraphics[width=1\textwidth]{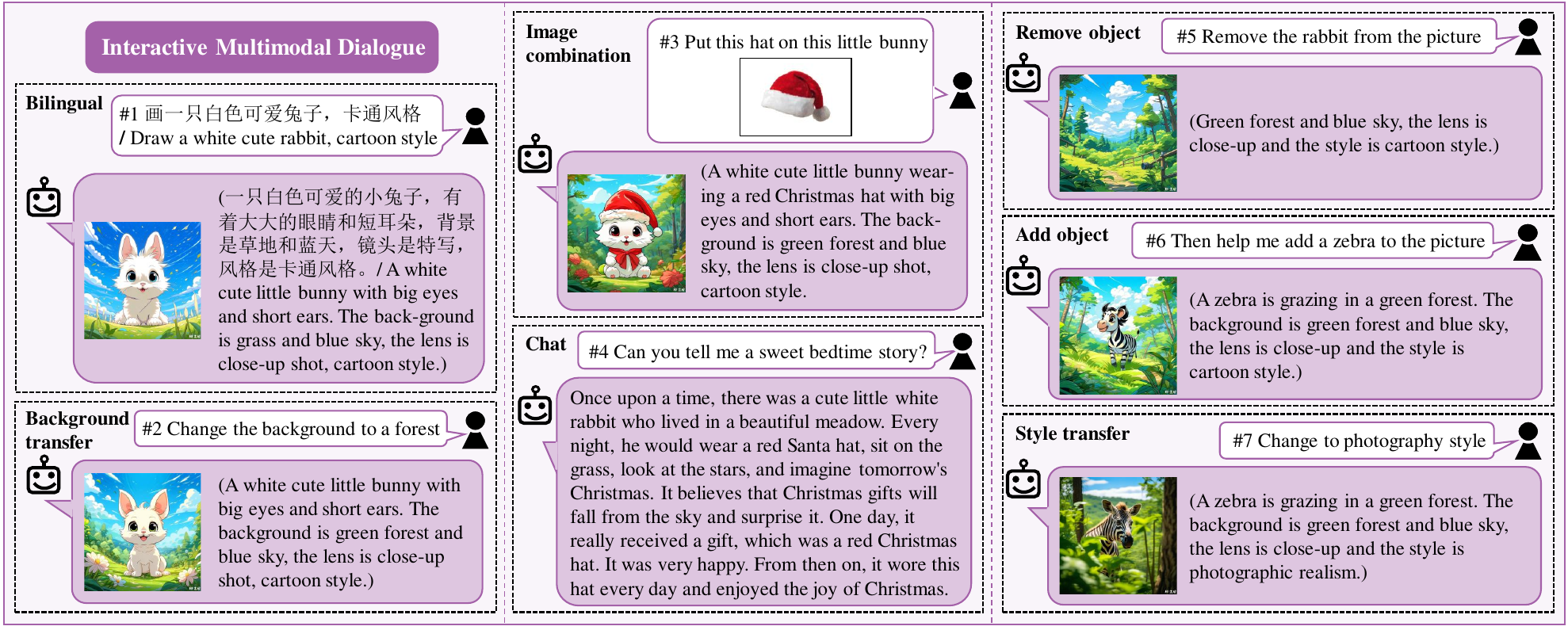}
    \captionof{figure}{Illustration of Multi-modal Interactive Dialogue System (MIDS) built by our proposed DialogGen that can perform multi-turn multi-modal tasks responding to user's natural language instructions to meet the users' needs for image generation, image editing, and chatting.}
\end{center}
\begin{abstract}
  Text-to-image (T2I) generation models have significantly advanced in recent years. However, effective interaction with these models is challenging for average users due to the need for specialized prompt engineering knowledge and the inability to perform multi-turn image generation, hindering a dynamic and iterative creation process. Recent attempts have tried to equip Multi-modal Large Language Models (MLLMs) with T2I models to bring the user's natural language instructions into reality. Hence, the output modality of MLLMs is extended, and the multi-turn generation quality of T2I models is enhanced thanks to the strong multi-modal comprehension ability of MLLMs. However, many of these works face challenges in identifying correct output modalities and generating coherent images accordingly as the number of output modalities increases and the conversations go deeper. Therefore, we propose \textbf{DialogGen}, an effective pipeline to align off-the-shelf MLLMs and T2I models to build a Multi-modal Interactive Dialogue System (MIDS) for multi-turn Text-to-Image generation. It is composed of drawing prompt alignment, careful training data curation, and error correction.
Moreover, as the field of MIDS flourishes, comprehensive benchmarks are urgently needed to evaluate MIDS fairly in terms of output modality correctness and multi-modal output coherence.
To address this issue, we introduce the Multi-modal Dialogue Benchmark (\textbf{DialogBen}), a comprehensive bilingual benchmark designed to assess the ability of MLLMs to generate accurate and coherent multi-modal content that supports image editing. It contains two evaluation metrics to measure the model's ability to switch modalities and the coherence of the output images.
Our extensive experiments on DialogBen and user study demonstrate the effectiveness of DialogGen in producing correct output modalities and coherent multi-modal outputs compared with other State-of-the-Art models. We hope that DialogBen can contribute to the community for building more powerful MIDS. 
% The code and data will be publicly available at \href{https://github.com/Centaurusalpha/DialogGen}{https://github.com/Centaurusalpha/DialogGen}.
The project page is \href{https://hunyuan-dialoggen.github.io/}{https://hunyuan-dialoggen.github.io/}.
\end{abstract}

\section{Introduction}
\begin{figure}[t!]
		\begin{center}
            \includegraphics[ width=1\linewidth]{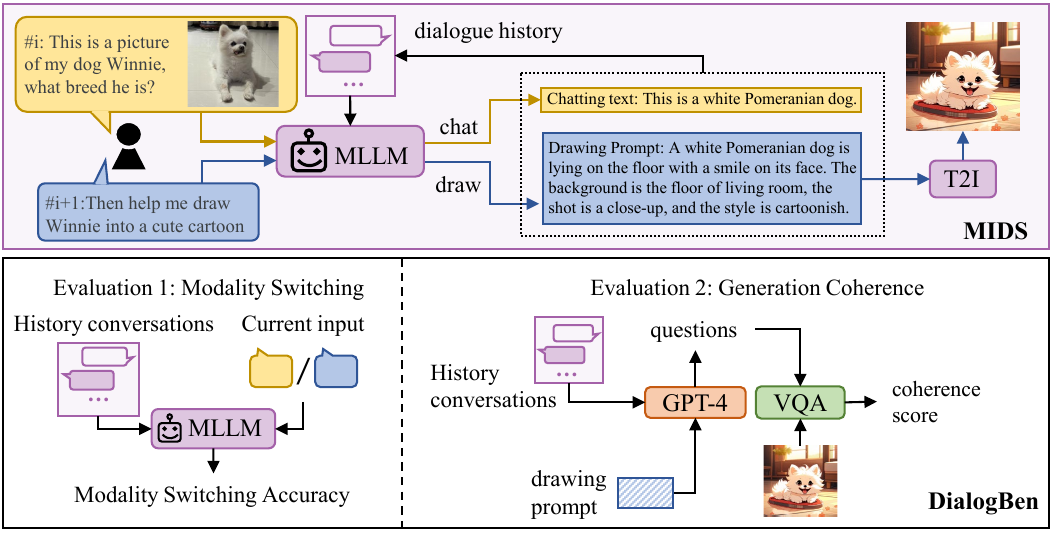}
		\end{center}
		\vspace{-2mm}
		\caption{
  % \hmb{modify the figure:remove the following score, emphasize the pipeline. }
  Overview of MIDS and DialogBen. MIDS can respond to the multi-modal user instructions with either a text response or a drawing prompt to be sent to a T2I model for image generation. DialogBen consists of 9957 three-turn multi-modal dialogs and two evaluation metrics to assess the capability of MIDS. 
  % \rh{What is the pseudo dialogue??? Need explanation? The icons of the dialogue history and the pseudo dialogue are the same, which seems to be confusing.}
  }
		\vspace{-5mm}
		\label{fig:pipeline}
\end{figure}

In recent years, Text-to-Image (T2I) diffusion models\cite{ho2020denoising,rombach2022high,song2020score,peebles2023scalable} emerges as a focal point of interest. These models, which generate visual content based on textual input, have the potential to revolutionize various industries, including advertising, entertainment, and education~\cite{zhang2017stackgan,el2019tell}. However, existing T2I models are still limited in comprehending user's complex natural language instructions due to the limited capacity of text encoders.
This leads to a big challenge for users when interacting with T2I models. Besides, prompt engineering expertise is required to accurately bring the user's instruction into reality, which hinders its broader application to average users.
Moreover, users may potentially generate images based on previous conversations and generation results, which calls for a multi-turn T2I generation. 
In response to this need, motivated by the strong language comprehension ability of Large Language Models (LLMs), an increasing number of researchers are turning their attention towards the integration of LLMs as a means to facilitate more natural and intuitive communication between users and T2I models. These works include Text-to-Image generation models equipped with open-source LLMs~\cite{wu2023next,ge2023making,koh2023generating,zeqiang2023mini,sun2023generative,koh2023grounding,chen2023llava} and closed-source LLMs such as DALLE3~\cite{betker2023improving} and RPG\cite{yang2024mastering}. This gives rise to a prosperous field of Multi-modal Interactive Dialogue System (MIDS) that receives multiple modalities and outputs multiple modalities, as can be seen at the top of Fig.~\ref{fig:pipeline}.
However, combining MLLMs and T2I models to build a powerful MIDS is non-trivial.
Experimental investigations reveal that contemporary solutions\cite{wu2023next,ge2023making} face considerable challenges in accurately identifying users' intentions and bringing the users' instructions into reality. This is particularly different from the success of those MLLMs that process inputs from diverse modalities and solely generate outputs in text format~\cite{chen2023llava,zhu2023minigpt,liu2023visual,zhang2023video,dai2023instructblip}.
If the output encompasses multiple modalities, it presents a significantly higher degree of complexity than single-modality output. Therefore, to build a more powerful MIDS that connects MLLMs and T2I models, we propose DialogGen, an effective pipeline to align off-the-shelf MLLMs and T2I models for building MIDS. We first conduct drawing prompt alignment that transforms the output of the MLLM to the form that T2I models favor. Later, we finetune the MLLM with carefully curated bilingual modality switching data. Besides, motivated by the learning process of a student who iterative learns and corrects his own mistakes guided by the teacher, we imitate the error correction process to further boost the performance of our MLLM, where the generation results are more aligned with human intention. Our DialogGen is compatible with any current State-Of-The-Art (SOTA) T2I models. 

\begin{figure}[b!]
		\begin{center}

\includegraphics[width=0.8\linewidth]{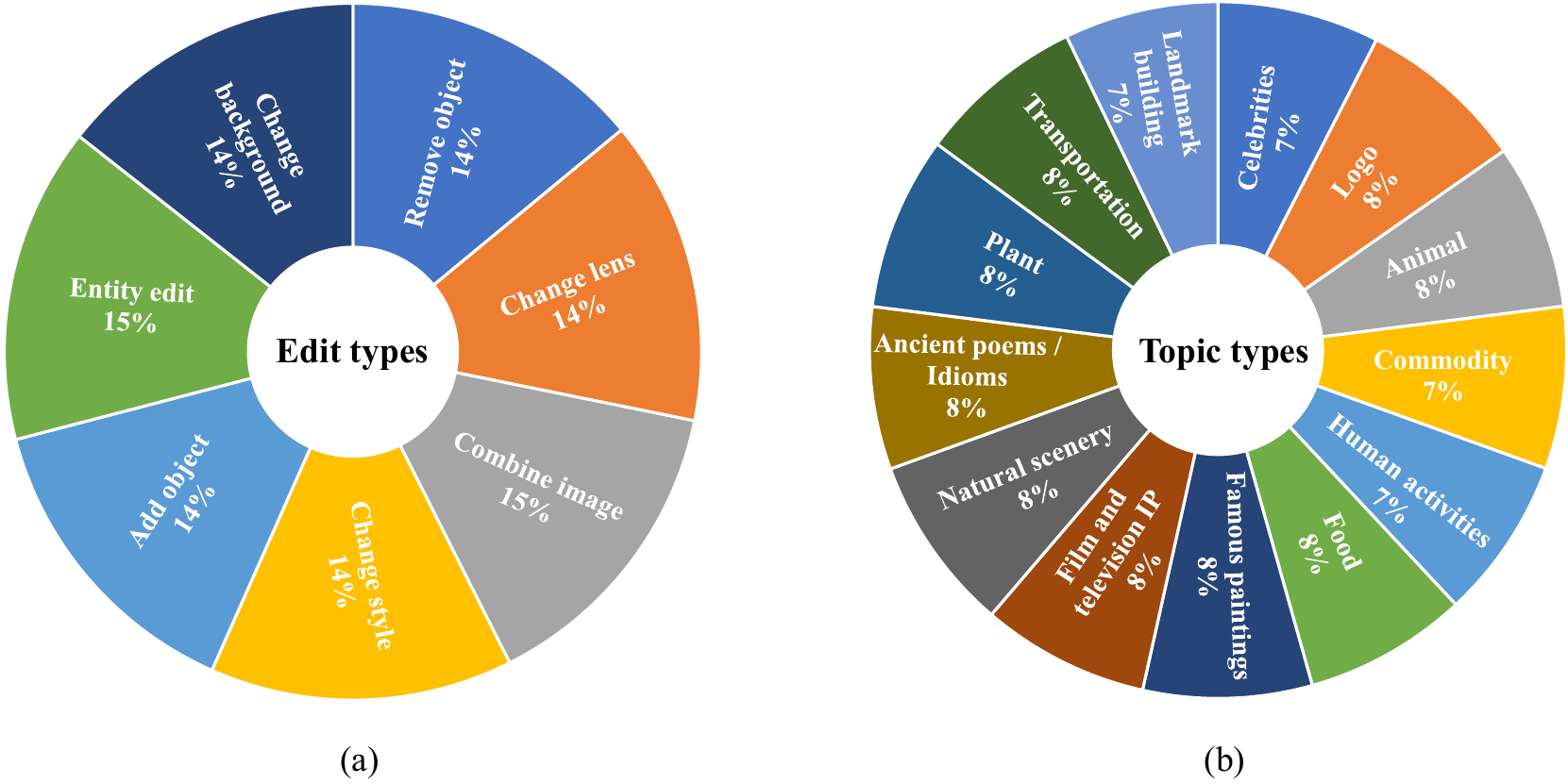}
% 			\fbox{\rule{0pt}{2in} \rule{.9\linewidth}{0pt}}
		\end{center}
		\vspace{-3mm}
		\caption{Our benchmark encompasses 7 edit instruction types and 13 topic types. 
  % \hmb{TBD: merge with other figures to save space}
  % \rh{The texts are too small!}\xd{add some figures otherwise too much blank}
  }
		\vspace{-4mm}
		\label{fig:intro_figure}
\end{figure}

As the field of MIDS continues to flourish, there is an urgent demand for a fair and comprehensive evaluation of such emerging systems. 
In response to this need, we introduce the Multi-modal dialogue Benchmark (DialogBen), an extensive benchmark to evaluate the capabilities of MIDS, which can engage in multi-turn dialogues with both input and output spanning multiple modalities. This benchmark contains the assessment of Modality Switching and Generation Coherence ability. It contains 9957 three-turn conversations covering 7 image editing types and 13 topic types, which is shown in Fig.~\ref{fig:intro_figure}. This benchmark aims to provide a proper assessment framework for the growing field of MIDS. By leveraging DialogBen, researchers can gain a deeper understanding of the challenges and opportunities in developing a more powerful dialogue system that can switch between different output modalities, generate semantically coherent images, and ultimately lead to more versatile and user-friendly MIDS. Fig~\ref{fig:DialogGen} illustrates the capabilities of MIDS built with our DialogGen. Comprehensive experiments on DialogBen have shown our superiority of DialogGen over current SOTA models.

To conclude, our contributions can be summarized as follows:
\begin{itemize}
\item We propose DialogGen, 
    % a seamless~\rh{``seamless'' is confusing}
    an effective pipeline to build an interactive multi-modal dialogue system for multi-turn Text-to-Image generation. It is built upon drawing prompt alignment, careful bilingual training data curation including object consistency guarantee, and instruction tuning data mixing. Besides, we also propose to train on error correction data from stronger LLMs to learn from mistakes it has made.
    \item We propose DialogBen, a comprehensive multi-turn bilingual benchmark that contains 9957 multi-turn multi-modal conversations covering 7 editing instruction types and 13 topic types to evaluate MIDS in terms of modality switching ability
    % , drawing instruction following
    , and the generation coherence. 
    \item Comprehensive experiments show that MIDS built by DialogGen beats current SOTA models in terms of modality switching accuracy, generation coherence VQA score, and user preference.
\end{itemize}

\section{Related Work}

\noindent\textbf{Text-to-Image Generation.}
Text-to-image (T2I) generation models, which accept natural language instructions and synthesize corresponding realistic images, have garnered significant attention at the intersection of computer vision and natural language processing. Notable approaches to T2I generation include the use of generative models, such as Variational Auto-encoders (VAEs) and Generative Adversarial Networks (GANs). The recent emergence of multi-modal models, such as CLIP~\cite{radford2021learning}, has further enhanced the alignment between image and text representations, leading to remarkable improvements in the capabilities of T2I models. These advancements have not only contributed to the growing interest in T2I generation but have also demonstrated the potential for further innovation and development in the field of multi-modal learning. Recently, diffusion models~\cite{song2020score,ho2020denoising,song2020denoising} achieved great success in producing high-fidelity images in a much more efficient and stable way, leading to the prosperity of T2I models.

\noindent\textbf{Prompting for Large Language Models.}
One of the most prominent strengths of Large Language Models (LLMs) is their in-context-learning\cite{dong2022survey,sun2023survey} ability. This enables users to effortlessly tailor LLMs for specific tasks or augment their capabilities using straightforward prompts. For instance, the chain-of-thoughts\cite{wei2022chain} technique pioneered the enhancement of LLMs by requesting them to produce a sequence of intermediate reasoning stages. Subsequently, various refined prompting methods have been developed, incorporating heuristics such as majority voting, backtracking, and graph-based thought representations. In our work, we rely on prompting GPT-4 for creating the benchmark dataset.

\noindent\textbf{Multi-modal Large Language Models and Evaluations.} Recent success of Large Language Models (LLMs)~\cite{touvron2023llama,touvron2023llama2,chowdhery2022palm,zhang2022opt} has largely benefited research on natural language processing. Inspired by the strong generation ability, researchers try to improve visual comprehension and Text-to-Image generation ability utilizing the generalization power of LLMs, leading to the prosperity of research on Multi-modal Large Language Models (MLLMs).
All MLLMs can receive inputs from multiple modalities; however, quite a few of them can produce the output from multiple modalities~\cite{wu2023next,ge2023making,chen2023llava,zeqiang2023mini,koh2023generating}, while others can only produce output in the text format.~\cite{zhu2023minigpt,liu2023visual,liu2023improved}. MLLMs that support image generation generally fall into two categories: one with the intermediate prompt for feeding into the image diffusion model, the other directly outputs the embedding vector for the diffusion model and generates the required images. Our DialogGen falls into the first category like DALLE3, as we believe that a user-friendly intermediate result is advantageous for analyzing the model behavior and providing feedback for model improvement.

The prosperous multi-modality-out MLLMs call for an effective benchmark to evaluate their capability. Our DialogBen specializes in the setting that users repeatedly seek image generation outputs rather than casual conversations. While previous literature such as MMDialog~\cite{feng2022mmdialog} provides a large-scale dataset for multi-turn multi-modal conversations, their focus is more on generating text-based scenarios. Moreover, they lack evaluations of the multi-turn image generation, because previous models often struggle to comprehend long conversations without the support of LLMs and MLLMs.

\section{DialogBen}
\label{sec:dialogben}
\subsection{Benchmark Design}
\label{subsec: bench design}
 The benchmark is designed to consist of a diverse set of dialogue scenarios that involve Text-to-Image generation. It encompasses various topics and image editing types, regarding different modality-switching scenarios for each turn of conversation. 
 To ensure broad coverage of conversational scenarios, we explore different combinations of input and output types based on four primary categories, i.e., $\textit{T}\to \textit{T}$, $\textit{T}\to \textit{I}$, $\textit{I+T}\to \textit{T}$ and $\textit{I+T}\to \textit{I}$, where $I,T$ represent image and text respectively. We also summarize the data proportion (see Table.7  in appendix) for each modality-switching scenario in each round and find that they are basically balanced. By selecting a type for each conversation turn, we devise $4^3=64$ three-turn conversation compositions.
 Subsequently, we employ GPT4 to generate `meta prompts' for these 64 types, each representing a unique conversational flow.
For each conversation combination, we traverse 13 predefined topics and 7 image editing methods (see Fig.~\ref{fig:intro_figure}) to yield approximately 15k samples after query GPT4 using corresponding `meta prompts'. 
To ensure the quality of the evaluation, we utilize the LLM model to filter out any mismatched modality output intentions between the input instructions and the ground truth label.
Later, we split an evaluation set that for each combination there is at least one sample and hence get an evaluation dataset comprising 9957 three-turn multi-modal conversations, which are characterized by incorporating both images and text as the input and output. While the remaining $\sim$ 4.5k samples are used as the training set. We refer to the training set and evaluation set as DialogBen-train and DialogBen-test.

\begin{minipage}{\textwidth}
\begin{minipage}[t]{0.49\textwidth}
  \begin{exmp}{Coherence VQA}{vqa}
    \begin{minipage}[t][3.5in][c]{\linewidth}
      \small
      \textbf{User}: 
      \includegraphics[width=0.4\textwidth]{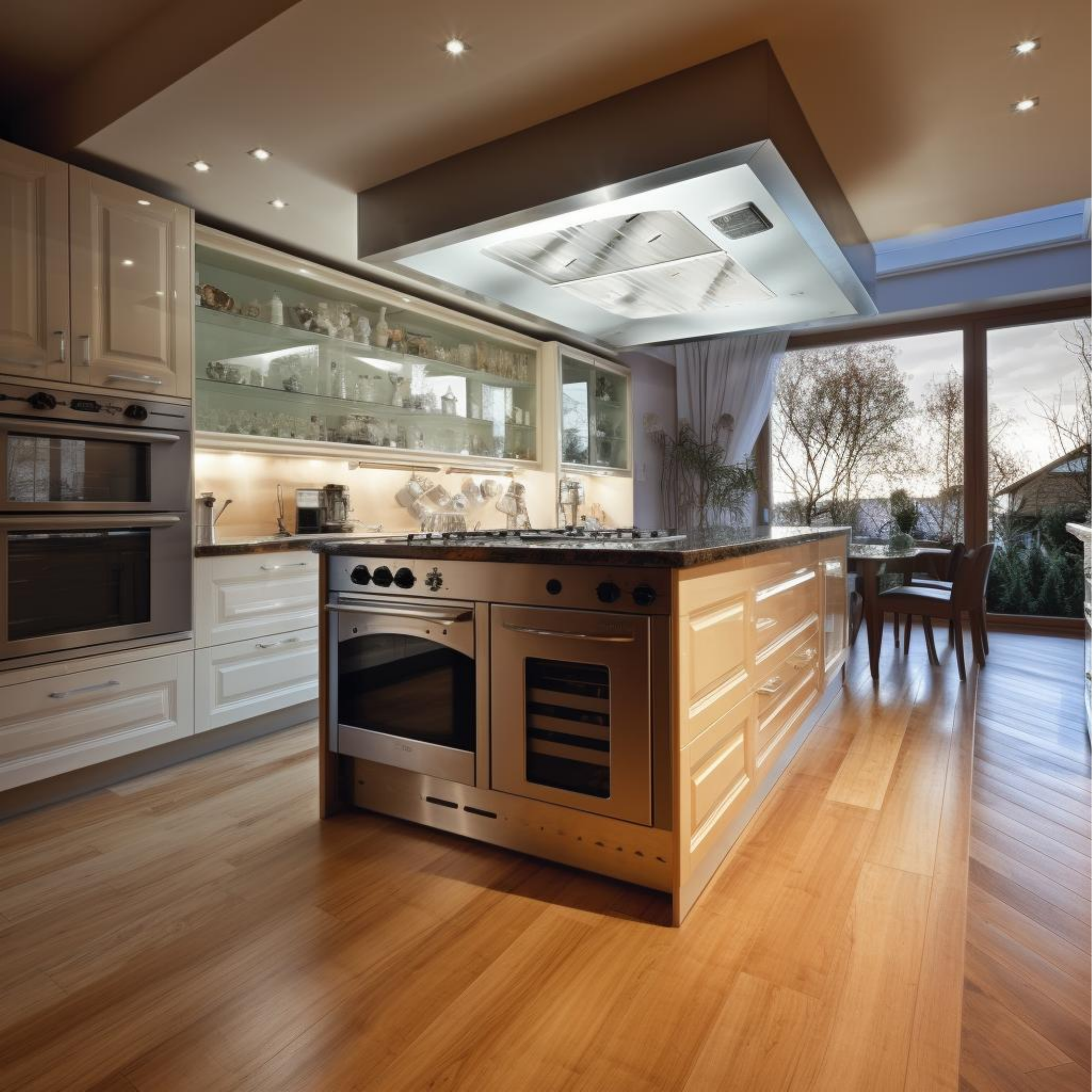}\\What kind of merchandise is this and what is it used for?
      \\
      \textbf{Assistant}: This product is an oven, mainly used for baking or roasting food, with temperature control and timing functions, and is a common appliance in modern home kitchens.\\
      \textbf{User}: What if you change the color of the oven so that it fits more in with the kitchen decor, say silver? \\
      \textbf{Questions: } \\
      1. Is there an oven in the picture? \\
      2. Is the oven placed in the middle of a kitchen? \\
      3. Is the background of the picture a modern kitchen? \\
      4. Is the shot of this picture panoramic? \\
      5. Is the color of the oven in the picture silver? \\
    \end{minipage}
  \end{exmp}
\end{minipage}
\hfill
\begin{minipage}[t]{0.49\textwidth}
  \begin{exmp}{Correction data}{correction_eg2}
    \begin{minipage}[t][3.5in][c]{\linewidth}
      \small
      \textbf{History:} (empty)\\
      
      \textbf{Question:} I'd like to know more about rockets. Could you draw me a picture of rockets?\\
      
      \textbf{Original Output:} A rocket is a large, powerful, and complex machine that is used to transport people and goods into space. It is usually made of steel and aluminum and is equipped with engines, fuel tanks, guidance systems, and other equipment. The picture shows a large rocket with a long tail, flying high in the sky.\\
      
      \textbf{Correction:} The output violates rule 3. The assistant's description misses the main point of the asking for a visual image of a rocket.\\
      
      \textbf{Correct Solution:} A rocket propelled upward by burning flames is moving through space, the Milky Way and stars in the background, the shot is panoramic, and the style is cartoonish.\\
    \end{minipage}
  \end{exmp}
\end{minipage}
\end{minipage}

\subsection{Evaluation Metrics}

\subsubsection{Modality Switching Accuracy}

Precisely identifying users' intentions and producing outputs in the suitable modality, whether with text or images, is essential for Multi-modal Interactive Dialog Systems (MIDS). In this way, the systems can deliver more relevant, informative, and contextually fitting responses. Therefore, we think it's important to check how well these systems can switch between modes of communication in our benchmark test.
We propose to evaluate the Modality Switching accuracy for each turn and focus on image and text modalities, as shown at the left bottom of Fig.~\ref{fig:pipeline}.
When a user inputs an image, it is usually accompanied by text describing the user's query related to the image. The system's output typically consists of image and text modalities. For each round, we assess 4 different modality switching scenarios mentioned in Sec.~\ref{subsec: bench design}, to ensure a comprehensive evaluation of the system's ability to generate the right multi-modal outputs. After inference on the benchmark, we calculate the modality switching accuracy as follows: suppose in the $i^{th}$ round, there are $n_{ij}$ samples belonging to $j^{th}$ modality switching scenario, let $c_{ij}^{(k)}$ be the binary variable, such that $c_{ij}^{(k)}=1$ indicates the model producing the correct output modalities. Then, the Modality Switching Accuracy $Acc_{ij}$ in the $i^{th}$ round for $j^{th}$ scenarios is calculated by
\begin{equation}
\label{equation: MS acc calculation}
    Acc_{ij} = 
    \left( \sum\limits_{k=1}^{n_{ij}}c_{ij}^{(k)} \right) / n_{ij}
    % \frac{\sum\limits_{k=1}^{n_{ij}}c_{ij}^{(k)}}{n_{ij}}
\end{equation}

% \vspace{-5mm}
\subsubsection{Generation Coherence}
To evaluate the generation quality in multi-turn conversations in image generation mode, and to better compare with MIDS that produce images directly without intermediate text prompts, we propose a Generation Coherence VQA score.

Inspired by \cite{huang2023t2i}, we employ the visual question answering (VQA) capability of BLIP\cite{li2023blip} to assess the generation quality.
As shown at the right bottom of Fig.~\ref{fig:pipeline}, for each query in our DialogBen-test, we utilize the historical conversation context, prompting GPT-4 to generate the required elements for the desired output images. These elements are then transformed into a set of alternative questions and their corresponding answers. This approach equips DialogBen-test with the affiliated collections of simple questions about the desirable output images.

During the evaluation, we employ the BLIP-VQA model~\cite{huang2023t2i} to process the images generated by candidate models and the alternative question. The probability of the correct answer to each question is used as the score. We then calculate the average score of all questions associated with the generated image to determine the image's score.
The Generation Coherence VQA Score for the entire test set is found by averaging these scores across all images.
Specifically, for each test query $q_i$ affiliated with $m_{i}$ questions and the generated image $I_i$, the generation coherence score $S_i$ is formulated as
\vspace{-3mm}
\begin{equation}
    S_i = Mean(\{P(BLIP(q_i^{(j)},I_i)=GT(q_i^{(j)},I_i))\}_{j=1}^{m_{i}}))
\end{equation}
\vspace{-3mm}

In Example~\ref{exmp:vqa} we showcase a sample from the candidate sets of the questions to be asked.

\section{DialogGen}
\begin{figure}[t!]
    \centering
    \includegraphics[width=\linewidth]{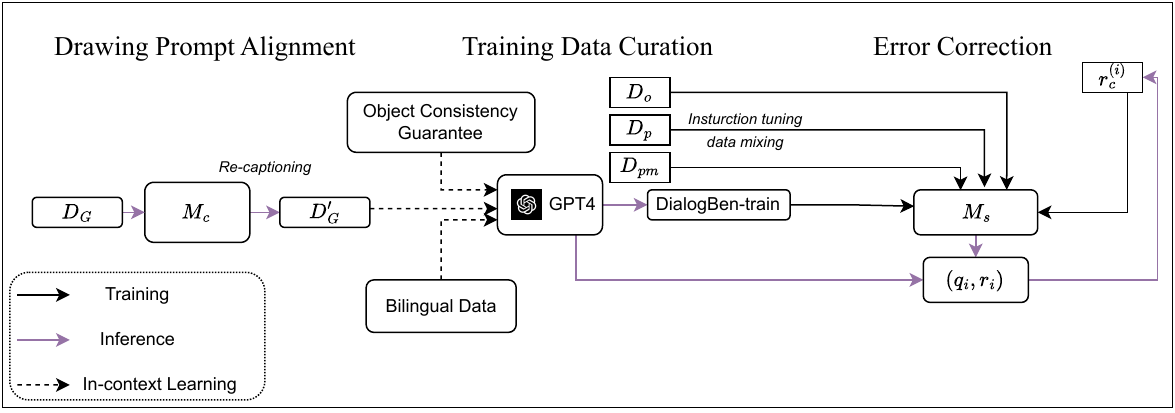}
    \caption{The overall pipeline of DialogGen which consists of Drawing Prompt Alignment, Training Data Curation, and Error Correction. In Drawing Prompt Alignment, re-captioning on $D_G$ is performed to ensure the alignment between transformed prompts and the T2I model. Then we carefully curate the training data such as adding object consistency guarantee, bilingual data and mixed instruction tuning data during training. Finally, we employ an error correction mechanism on student model $M_s$ to make the model learn from its mistakes.}
    \vspace{-5mm}
    \label{fig:DialogGen-pipeline}
\end{figure}

In this section, we introduce the components of DialogGen, which include Drawing Prompt Alignment, Training Data Curation, and Error Correction. Given an off-the-shelf MLLM $M_C(I,T)$ and a T2I model $G$, DialogGen integrates them seamlessly to build a powerful MIDS. The overall framework can be seen in Fig.~\ref{fig:DialogGen-pipeline}. We will introduce them parts by parts.

\subsection{Drawing Prompt Alignment}
Suppose the original training data of $G$ is $D_G = (I_i,T_i) \sim P(I,T)$, where $I$ and $T$ represent image and text, respectively. The user's text instructions should be aligned with the distribution of $D_G$ for optimal image generation. Yet, because $D_G$ typically comes from various data resources, it is hard to get the ground truth of $P$.

However, since re-captioning has been shown effective in various works\cite{betker2023improving,yang2024mastering} for improving better T2I generation quality, we utilize a re-caption model for text refinement. For a given MLLM $M_C$, we first apply it to re-caption the training set of $G$ to obtain a new training set $D_G'=(I_i,T_i') \sim P'(I,T)$, where $T_i'=M_C(I_i,P_{cap})$ and $P_{cap}$ is the caption prompt such as "Please describe this image in detail.". 
Then we re-train the T2I model on $D_G'$. Afterward, when querying GPT-4 for training data generation, we sample $(I'_{icl},T'_{icl}) \sim P'(I,T)$ as the in-context learning samples in the ``meta-prompts'', aligning the generated prompts with $P'$. Thus, the training data on image generation would have a similar drawing prompt distribution as $P'$, which is exactly what $G$ is re-trained on. 
In this way, the transformed drawing prompts match the given T2I models and thus lead to the expected image generation results.
If we could directly access the ground truth of $P$ without re-captioning through $M_C$, we would simply draw samples $(I_{icl},T_{icl})$ from $P(I,T)$ for in-context learning to guide data generation process. 

\subsection{Training Data Curation}
\label{sec:4.2}
% \hmb{@yanxin} 
\noindent\textbf{Object Consistency Guarantee.}
In multi-turn image generation, our goal is to ensure that objects generated across different conversational turns remain as consistent as possible. 
To achieve this, we add the following constraints for training data generation: for image generation that builds upon the images produced in previous rounds, the transformed drawing prompts should satisfy the user's current demand while being altered as little as possible from the drawing prompts used for previous images.
Fig.\ref{fig:DialogGen} \#6 and \#7 show the minor differences between the prompts in consecutive turns.
Moreover, during the inference phase of a given conversation, we fix the random seed of the T2I model. 
This approach significantly increases the likelihood of maintaining object consistency throughout the dialogue.

\noindent\textbf{Instruction Tuning Data Mixing.}
To maintain the multi-modal conversation ability, we’ve included a range of open-sourced uni/multi-modal conversation datasets, denoted as $D_o$.
Besides, to address the lack of datasets that turn the user's natural language input into detailed description prompts that fall into 
$P'$, the suitable distribution of $D_G$ for $G$ to generate images, We first collect $n=10$ manually-annotated labels.
These labels will transform user ideas into semantically appropriate, detailed prompts that align with $P'$.
Using these labels as in-context samples, we then query GPT-4 to generate a dataset that contains single-turn text-to-prompt samples, referred to as $D_p$. Since we aim at building a multi-turn dialogue system, we randomly shuffle and concatenate the single-turn samples from $D_o$ and $D_p$ to get a pseudo-multi-turn dataset $D_{pm}$.
This dataset features multi-turn conversations not necessarily preserving semantic coherence,
simulating the scenarios in which the user may switch the topic within a conversation. More importantly, we mix the collection of $D_o$, $D_p$, $D_{pm}$ together with $D_T$, the training set of DialogBen. Please refer to Table.5 in the appendix for detailed compositions of the instruction tuning data and training data samples.

\noindent\textbf{Bilingual training}
Current MIDS typically only supports English scenarios and is trained on English data.
Motivated by work~\cite{chen2023breaking} that multilingual training can benefit LLMs' reasoning ability, we extend the training data from unilingual to bilingual.
Specifically, we develop a Chinese counterpart of DialogGen, which is
identical in structure to the original but differing in language.
We refer them to DialogBen-en and DialogBen-cn, respectively. 
We conduct experiments on these versions individually as well as their mixed version. Please refer to Sec.~\ref{sec:exp} for more details.

\begin{table*}[ht!]
% \renewcommand\arraystretch{0.8}
% \Huge
\caption{Comparison of Modality Switching (MS) accuracy (\%) of different models under standard fine-tuning and error correction counterparts with different style.
}
\label{tab:correction_exps}
\centering
\resizebox{.8\linewidth}{!}{
\begin{tabular}{@{}clccc@{}}
\toprule
\multirow{2}{*}{Model} & \multicolumn{1}{c}{\multirow{2}{*}{Training}} & \multirow{2}{*}{\#Inference}& \multicolumn{2}{c}{DialogBen-MS} \\ \cmidrule(lr){4-5}
 & & & 1st / 2nd / 3rd & Avg.  \\ \midrule
\multirow{3}{*}{DialogGen-A} & Standard Fine-Tuning & 1-step &  93.6 / 87.1 / 84.6 & 88.4   \\
 & \cellcolor{purple1!25} + Error Correction & \cellcolor{purple1!25}1-step & \cellcolor{purple1!25}96.9 / 95.6 / 93.4 & \cellcolor{purple1!25}95.3(+6.9)    \\
  & \cellcolor{purple1!25} + Error Correction & \cellcolor{purple1!25}2-step & \cellcolor{purple1!25} 97.8 / 97.4 / 93.6 & \cellcolor{purple1!25}96.2(+7.8) 
  \\ \midrule
\multirow{3}{*}{DialogBen} & Standard Fine-Tuning & 1-step & 96.4 / 96.5 / 95.6 &  96.1\\
 & \cellcolor{purple1!25} + Error Correction  &  \cellcolor{purple1!25}1-step & \cellcolor{purple1!25}97.1 / 97.9 / 96.7  & \cellcolor{purple1!25} 97.2 (+1.1)  \\
 & \cellcolor{purple1!25} + Error Correction & \cellcolor{purple1!25}2-step & \cellcolor{purple1!25} 97.3 / 97.9 / 96.9 &   \cellcolor{purple1!25}97.4 (+1.3)  
 \\ \bottomrule
 \vspace{-6mm}
\end{tabular}
}
\end{table*}

\subsection{Error Correction}

\noindent\textbf{Data Collection.}
To enhance our models' understanding of users' intentions, collecting more data is a common but often expensive approach.
Instead, we propose learning from previous models' errors as a more efficient strategy.
Specifically,
given a training dataset $D$, we obtain a student model $M_s$ after several training epochs. We then inference
$M_s$ on $D$ to generate a collection of potentially noisy query-response pairs $(q_i,r_i),i=1,2,...,|D|.$
Next, we introduce a teacher corrector model $M_T$ and a prompt $P_f$ to query $M_T$, where $P_f$ is the few-show prompt to instruct $M_T$ to generate error correction data, as detailed in the Appendix.
We employ the corrector model $M_T$ to generate a correction $c_i$ for each sample in the training set:
\begin{equation}\label{sec:sample_correction}
    c_{i} \sim \mathcal{M}_T (P_f \oplus q_{i} \oplus r_{i})
\end{equation}

Note that in $P_f$ we set up 3 rules for $M_T$ to determine whether the output is satisfactory. Therefore, we include 3 human annotated error-correction examples as the few-shot in-context learning samples. In Example \ref{exmp:correction_eg2}, there is an example showcasing the correction data generated by $M_c$.

\noindent\textbf{Training.} 
During training, we use a correction prompt $P_c$ to prompt the student model to obtain the correction response $r_c^{(i)}$,
\begin{equation}
    r_{c}^{(i)} \sim M_s (P_c \oplus q_{i} \oplus r_{i})
\end{equation}

The training loss is calculated between both the correction response and the provided suitable output.
Note that we include both the feedback from correct and incorrect responses. Feedback from correct responses enhances the model's confidence in its accurate actions. On the other hand, feedback from incorrect responses alerts the model to its mistakes and how to avoid them in the future.
By incorporating this error correction mechanism into MLLM, we would create a more robust and reliable system capable of generating high-quality, contextually appropriate responses.
This approach not only helps the model to learn from its mistakes but also fosters a deeper understanding of the nuances and complexities involved in multi-modal interactive dialogue tasks.

\begin{table*}[t]
% \renewcommand\arraystretch{0.8}
% \Huge
\caption{Comparison of adding bilingual training on Modality Switching (MS) Accuracy (\%) of models under different training data. \textit{cn} and \textit{en} represent Chinese and English data respectively.
}
\label{tab:multilingual}
\centering
\resizebox{.99\linewidth}{!}{
\begin{tabular}{@{}clccccc@{}}
\toprule
\multirow{2}{*}{Model} & \multicolumn{1}{c}{\multirow{2}{*}{Training}} & \multicolumn{1}{c}{\multirow{2}{*}{Evaluation Data}} & \multicolumn{4}{c}{DialogBen-MS}  \\   \cmidrule(l){4-7} 
 & \multicolumn{1}{c}{} &  & Round1 & Round2 & Round3 & Avg  \\ \midrule
\multirow{2}{*}{Qwen-VL} & DialogBen-train-$\textit{en}$ &DialogBen-test-\textit{en}  & 77.8 / 97.6 / 97.0 / 95.6 & 77.5 / 98.4 / 96.2 / 88.7 & 79.9 / 95.6 / 95.7 / 79.6 & 90.0  \\
 & \cellcolor{purple1!25} + DialogBen-train-\textit{cn} & DialogBen-test-\textit{en}  & \cellcolor{purple1!25}78.2 / 96.9 / 97.0 / 96.0 & \cellcolor{purple1!25}83.0 / 97.6 / 96.6 / 89.5 & \cellcolor{purple1!25}82.5 / 96.7 / 95.4 / 85.4 & \cellcolor{purple1!25}91.3 (+1.3)  \\ \midrule
\multirow{2}{*}{Qwen-VL} & DialogBen-train-\textit{cn}  & DialogBen-test-\textit{cn}  & 82.5 / 98.0 / 97.0 / 97.0 & 76.0 / 99.0 / 94.0 /92.0 & 71.0 / 97.9 / 93.6 / 83.8 & 90.1  \\
 & \cellcolor{purple1!25} + DialogBen-train-\textit{en} & DialogBen-test-\textit{cn}   & \cellcolor{purple1!25}87.0 / 98.4 / 97.7 / 98.7 & \cellcolor{purple1!25}90.4 / 99.2 / 95.3 / 95.0 & \cellcolor{purple1!25}86.0 / 98.0 / 92.6 / 89.2 & \cellcolor{purple1!25}94.0(+3.9)  
 \\ \bottomrule
\end{tabular}
}
\end{table*}
\section{Experiments}
\label{sec:exp}
\subsection{Models and Baselines}
\noindent\textbf{Base model.}
Note that we aim to build a strong bilingual Mutli-modal Interactive Dialogue System, so the base MLLM model needs to have a considerable perception ability in both Chinese and English. Therefore, we carefully choose our base MLLM model as Qwen-VL~\cite{bai2023qwen}, which is an open-sourced Multi-modal Large Language Model that supports both Chinese and English input and output. It is capable of receiving images and texts as input and performing Multi-modal comprehension tasks like text reading, and image understanding in the format of text. For T2I models, we conduct experiments on Hunyuan T2I\footnote{https://hunyuan.tencent.com/}, which is a closed-source T2I model of ours. This is because the re-captioning is performed on the training dataset of Hunyuan T2I. For fair comparison in Coherence VQA score, we also conduct experiments for Stable Diffusion v1.5 and Stable Diffusion v2.1.

\noindent\textbf{Baselines.}
\begin{enumerate}
    \item Qwen-VL-few-shot: We provide a simplest way to equip Qwen-VL with T2I model $G$ without training. We first add the description of the system we aim to build to serve as the system prompt during inference. Then we provide $n$ manually designed samples demonstrating how to transform the multi-turn user instruction into drawing prompts to be sent to $G$, where $n=0,1$.
    \item NExT-GPT\cite{wu2023next}: It is one of the earliest works to introduce the concept of any-to-any MLLMs. They introduce a projection layer to align the X-encoder (X can be image, audio, or video) and X-diffusion model with the LLM embedding space to build an MIDS that receives text, image, audio, and video both as candidate input and candidate output.
    \item SEED-LLaMA\cite{ge2023making}: It designs a SEED tokenizer to tokenize the image and train the multi-modal discrete sequence in a unified objective. Therefore, when it needs to perform image generation, image tokens can be predicted autoregressively during inference and hence be decoded into images. 
\end{enumerate}

\subsection{Training and Inference}
To make it compatible with our goal of performing interactive multi-modal image generation, we add a modality switch token \textless{}draw\textgreater{}
before the model output to identify the recognized intention of the user and a training prompt $P_T$ as "Please first identify the intention of the user, if it is to draw please append \textless{}draw\textgreater{} before the output". 

To emphasize the effectiveness of $D_T$, we named the refer DialogGen-A as the model trained with $D_A=D_o \cup D_p \cup D_{pm}$, while DialogGen-B is the model trained with 
$D_B=D_o \cup D_p \cup D_{pm} \cup D_T$.

For the Error Correction Data, we use GPT-4 as the $M_T$, which is the most powerful closed-source LLM. We first generate $n=100$ Error Correction Data and evaluate the quality by humans. We found that 98\% of them give right correction about the user's intention and 85\% of them give the right correction about transforming the user's needs into correct drawing prompts. Therefore, we consider the quality of the error correction to be acceptable for further training.

During inference, given a well-trained model $M_w$, we apply the same prompt $P_I$ as training. That is, $P_I=P_T$.

\noindent\textbf{Two-Step Inference}
To fully explore our model's ability, we additionally introduce a two-step inference procedure given that our model is fine-tuned with error correction data to be blessed with the ability to conduct self-correction. To be more specifically, given the query of the $i^{th}$ test example $q_t^{(i)}$, we first use $P_i$ to get a response $r_1^{(i)}$, where 
\begin{equation}\label{two step inference}
    r_1^{(i)} \sim M_w (P_I \oplus q_{t}^{(i)} )
\end{equation}

Afterwards, $r_1^{(i)}$ is sent into the model again with $q_t^{(i)}$ together with the correction prompt $P_c$ and get the corresponding correction $EC^{(i)}$ and suitable output $r_2^{(i)}$. We choose $r_2^{(i)}$ to be the final output.
\begin{equation}\label{two step inference}
    EC^{(i)}, r_2^{(i)} \sim M_w (P_c \oplus q_{t}^{(i)} \oplus r_1^{(i)} )
\end{equation}

\subsection{Evaluation}

\noindent\textbf{Modality Switching.}
\begin{table*}[t]
% \vspace{-5pt}

\vspace{5pt}
\centering
    \caption{Benchmarking on Modality Switching ability of different models, with T and I represents text and image respectively. On the left of $\rightarrow$ is the input modality, while on the right of  $\rightarrow$ is the output modality. The accuracy is calcualted by Equation.\ref{equation: MS acc calculation}. The best result is emphasized in \textbf{bold}. } 
\label{tab:intention_results}
% \vspace{0pt}

\resizebox{\linewidth}{!}{%  
\begin{tabular}{lccc>{}cc>{}cccc>{}cc>{}cc}
\toprule
\multicolumn{1}{c}{\multirow{2}{*}{\textbf{Model}}} & \multicolumn{4}{c}{\textbf{Round1}}   & \multicolumn{4}{c}{\textbf{Round2}} &    \multicolumn{4}{c}{\textbf{Round3}} & \multirow{2}{*}{\textbf{Avg.}}    \\
\cmidrule(lr){2-5}\cmidrule(lr){6-9}\cmidrule(lr){10-13}
\multicolumn{1}{c}{}    & $\textit{T}\to \textit{T}$ & $\textit{T}\to \textit{I}$ & $\textit{I+T}\to \textit{I}$ & $\textit{I+T}\to \textit{T}$ & $\textit{T}\to \textit{T}$ & $\textit{T}\to \textit{I}$ & $\textit{I+T}\to \textit{I}$ & $\textit{I+T}\to \textit{T}$ & $\textit{T}\to \textit{T}$ & $\textit{T}\to \textit{I}$ & $\textit{I+T}\to \textit{I}$ & $\textit{I+T}\to \textit{T}$ \\
              \midrule
Qwen-VL~\cite{bai2023qwen}-0-shot & 92.46 & 7.42 & 1.11 & \textbf{100} & 76.75 & 34.13 & 2.54 & \textbf{100} & 78.16 & 27.98 & 4.24 & 99.58  & 52.03 \\
Qwen-VL~\cite{bai2023qwen}-1-shot & 93.90 & 8.42 & 1.04 & \textbf{100} & 79.78 & 34.27 & 2.29 & \textbf{100} & 79.48 & 27.58 & 3.98 & \textbf{100} & 52.56   \\
NExT-GPT~\cite{wu2023next} & 74.89 & 59.37 & 58.88 & 90.43 & 89.66 & 38.95 & 31.35 & 91.21 & 96.06 & 16.87 & 20.49 & 93.75 & 63.49  \\
SEED-LLaMA~\cite{ge2023making} & 94.71 & 89.69 & 92.30 & 97.18 & 98.79 & 84.82 & 85.98 & 98.13 & \textbf{99.01} & 74.77 & 76.74 & 97.28 & 90.78\\
% SEED-LLaMA~\cite{ge2023making}     \\
DialogGen-A & 80.00 & \textbf{97.66} & 97.04 & 95.25 & 65.56 & 98.40 & 93.64 & 88.75 & 64.93 & \textbf{98.77} & 93.28 & 76.85 & 87.51    \\
DialogGen-B & 85.41 & 98.44 & 98.15 & 97.87 & 90.47 & \textbf{99.20} & \textbf{99.57} & 94.58 & 93.93 & 100 & 99.00 & 91.00 & 95.63 \\

% LLaVA-interactive~\cite{chen2023llava}  & \\
% GILL~\cite{koh2023generating}  & \\

% \abbr-A (ours)        & 0.3390 & 0.7667 & 0.8219 & 0.6414   & ?    & ? & 0.3175 & 0.7149 & 0.7630 & 0.4546   & ?    & ? \\
\midrule

DialogGen & \textbf{94.97} & 96.87 & \textbf{98.88} & 97.82 & \textbf{97.04} & 98.79 & \textbf{99.00} & 95.81 & 95.19 & 97.53 & 98.23 & 95.83 & \textbf{97.24}  \\
              \bottomrule
\end{tabular}
}
\vspace{-3mm}
% \vspace{0pt}
% \vspace{5pt}
\end{table*}
In Table~\ref{tab:intention_results} we include the comparison result on Modality Switching accuracy of different models.
Note that both Qwen-VL-0-shot and Qwen-VL-1-shot have poor performance in identifying the user's drawing intention, which is reflected in the small number in $\textit{I+T}\to \textit{I}$. We also check the inference output for $\textit{I+T}\to \textit{T}$ and find that these two models use texts to reject to answer the question related to Visual-Question-Answering in most of the cases, leading to the intention being recognized as correct. This contributes to their near-perfect performance in scenario $\textit{I+T}\to \textit{T}$.

As can be seen in the table, for DialogGen-A, the model tends to recognize the user's intention as draw rather than talk and hence has relatively high accuracy when the output modality is image. For NExT-GPT, it has higher accuracy in the case when the output modality is text and severely worse performance when the output modality is image. This is because it has four output modalities. When the ground truth modality is image there are often hints in the query to inform the model to produce content other than text, which somehow lets the model produce audio or video instead.
And as the conversations go deeper, both the two models show several performance degradations in the scenario that they are not good at.

For experiments including error correction (Table~\ref{tab:correction_exps}), we can see that by incorporating Error Correction data, both models can have obvious performance improvement, especially when the diversity of the training dataset is relatively small (DialogGen-A), the average success rate of switching modalities increases from 88.4 to 95.3. Besides, incorporating error correction data helps improve the stability of the model as the conversations go deeper. As for the 2-step inference, the benefit is less obvious.

For comparative experiments including bilingual training (Table~\ref{tab:multilingual}). To better illustrate the effectiveness of adding bilingual data, we eliminate all the training except DialogBen-train and split it into English (DialogBen-train-en) and Chinese (DialogBen-train-cn). We can figure out that both the model trained in English and the model trained in Chinese can benefit from adding the same training data except for the language difference.

\begin{figure}[h]
		\begin{center}
            \includegraphics[ width=1\linewidth]{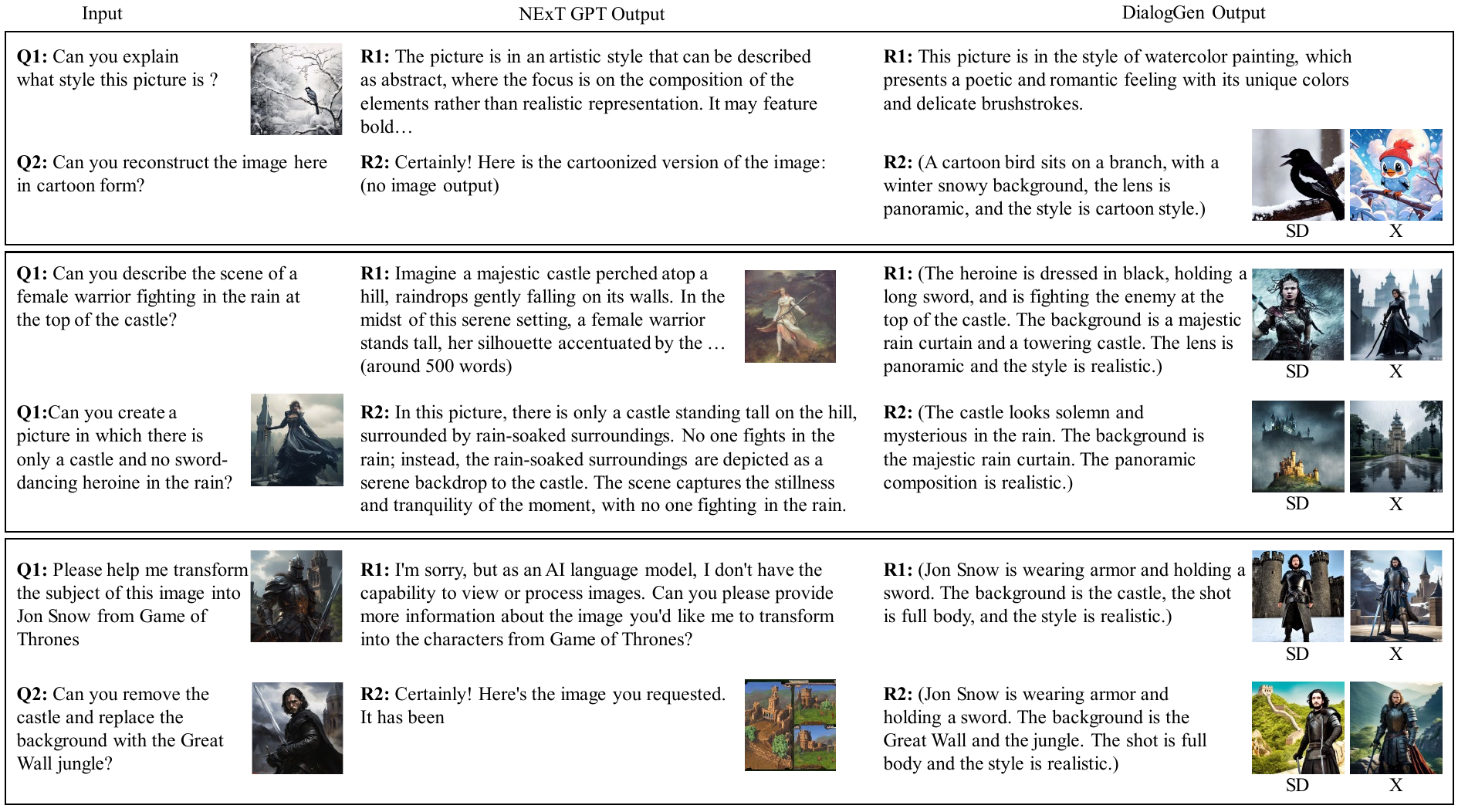}
		\end{center}
		% \vspace{-4mm}
		\caption{Visualization of output results of NexTGPT, DialogGen-SD, and DialogGen-Hunyuan on the DialogBen benchmark. DialogGen has better performance in generating output of correct modality and higher semantic coherence.}
		% \vspace{-4mm}
		\label{fig:vis}
\end{figure}

\noindent\textbf{Coherence VQA.}
Note that our DialogGen builds a multi-modal agent that connects with users and T2I models and hence is compatible with any T2I models. We provide the coherence VQA score and the human evaluation\footnote{Details can be found in Appdendix.} of different Multi-modal Interactive Dialogue systems.(See Table~\ref{tab: coherence_vqa}) Here DialogGen-Hunyuan is DialogGen equipped with a closed-source model Hunyuan T2I and DialogGen-SD-v1-5 and DialogGen-SD-v2-1 is DialogGen equipped with Stable Diffusion~\cite{rombach2022high} v1.5 and v2.1 respectively. They are the same T2I models equipped with NExT-GPT and SEED-LLaMA.
\begin{table}[t!]
\caption{Comparison of Cohence VQA Score on different models. $^{\uparrow}$ indicates higher is better, \textbf{bold} indicates the best result. 
DialogGen-X has the best performance and the Coherence VQA Score perfectly aligns with Human evaluation.
}
% \vspace{-2mm}
\small
    \setlength{\tabcolsep}{12pt}

    \centering
    \resizebox{.85\linewidth}{!}{
    \begin{tabular}{l|cc}
    \toprule
           Model  & Coherence VQA Score$^{\uparrow}$ & Human$^{\uparrow}$   \\
           \midrule
           NextGPT-SD-v1-5& 0.5153 & 0.5524\\
           DialogGen-A-SD-v1-5 & 0.6371  & 0.6732 \\
         DialogGen-SD-v1-5  & 0.6403  & 0.6911 \\
            SEED-LLaMA-SD-v2-1 & 0.5776 & 0.6313 \\
           DialogGen-SD-v2-1 & 0.6468  & 0.7277 \\
        DialogGen-X  & \textbf{0.6514} & \textbf{0.7559}\\
    \bottomrule
    \end{tabular}
    }
    % \vspace{-3mm}
    
    % \vspace{-5mm}
    \label{tab: coherence_vqa}
\end{table}

\begin{figure}
    \centering
    \includegraphics[ width=0.95\linewidth]{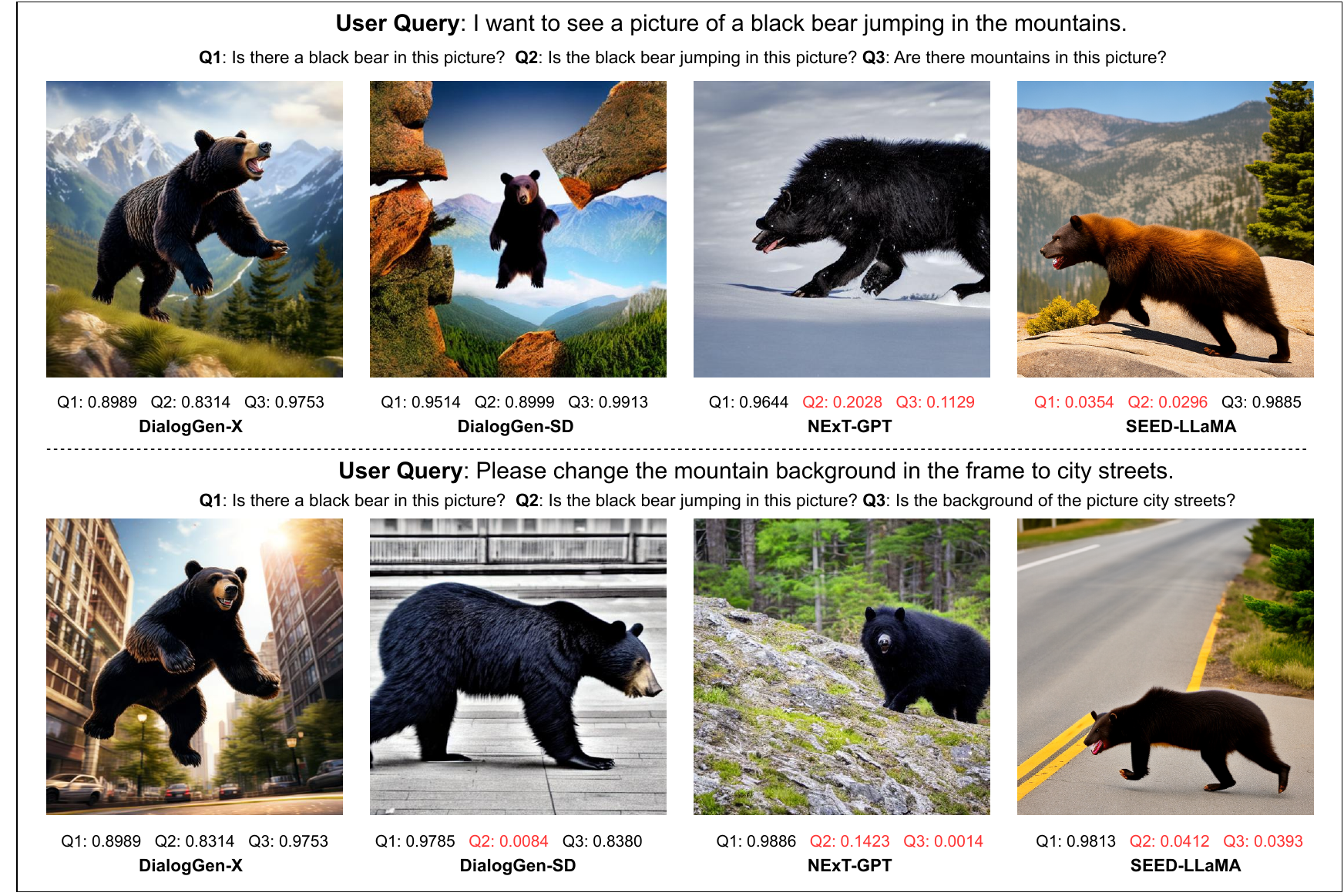}
    \caption{Visualization of Coherence VQA on different models. The numbers after each question are the probability of the model answering correctly about the corresponding question. \textcolor{red}{Red} numbers indicates the wrong answers. }
    \vspace{-5mm}
    \label{fig:vqa_result}
\end{figure}

In Fig~\ref{fig:vqa_result} we provide the visualization of the Coherence VQA Score of different models. 
As can be seen in the figure, in the first turn of the dialogue, both DialogGen-Hunyuan and DialogGen-SD generate correct images except for a flaw in the action \textbf{jumping} for DialogGen-SD. While both NExT-GPT and SEED-LLaMA make two mistakes and they are revealed by the low coherence VQA score. In the second turn of the dialogue, DialogGen-SD misses the action \textbf{jumping} and both NExT-GPT and SEED-LLaMA make two mistakes. 
Looking more closely we can find the DialogGen-Hunyuan has a relatively better consistent object, which implies the effectiveness of Object Consistency Guarantee in Sec.\ref{sec:4.2}. Besides, given that the coherence VQA score perfectly aligns with human evaluation, we consider the coherence VQA score to be a suitable proxy that reflects what extent the output image satisfies the user's need.

\noindent\textbf{Qualitative Examples.}
In Fig.~\ref{fig:vis}, we present some qualitative examples of our model and NExT-GPT. All the Q1 and Q2 are coherent conversations. The content within the bracket is transformed drawing prompt for the given scenario. Our model generates correct drawing prompts in response to the user's demand in all these cases. Noticeably, on the top right R2 the correct prompt failed to let Stable Diffusion generate the correct style while ours (X) succeeded. This is because our DialogBen-train is collected so that it has already been aligned with X.
We refer the readers to the Appendix for more qualitative examples and demo conversations. We will also release a demo soon.

\section{Conclusions and Discussions}
\label{sec:conclusion}
In this work, we propose DialogGen, a Multi-modal Interactive Dialogue system (MIDS) for Multi-turn Text-to-Image generation, which is trained with carefully curated instruction tuning and Error Correction data generated by more powerful LLMs. In addition to that, in response to the need for a comprehensive benchmark to evaluate rapidly emerging new MIDS, we propose DialogBen. It consists of 9957 three-turn multi-modal conversations together with two evaluation metrics, namely modality switching accuracy and coherence VQA score to comprehensively assess the ability of a Multi-modal Interactive Dialogue System. 

\noindent\textbf{Limitation}
The limitation lies in the resource requirement when conducting the re-captioning for the training data of the T2I model. However, fine-tuning open-sourced T2I models with a small subset of high-quality data with known distribution $P_{hq}$ has been widely adopted. In this case, we can simply sample $(I_{icl},T_{icl}) \sim P_{hq}(I,T)$ to simplify the procedure.

\noindent\textbf{Future work}. Current training data collection is aligned with the T2I model's characteristics. It can be designed to be aligned with human preference in the future and hence we can train the MLLMs to satisfy human preference using algorithms like\cite{rafailov2024direct}

% {
% \small

% [1] Alexander, J.A.\ \& Mozer, M.C.\ (1995) Template-based algorithms for
% connectionist rule extraction. In G.\ Tesauro, D.S.\ Touretzky and T.K.\ Leen
% (eds.), {\it Advances in Neural Information Processing Systems 7},
% pp.\ 609--616. Cambridge, MA: MIT Press.

% [2] Bower, J.M.\ \& Beeman, D.\ (1995) {\it The Book of GENESIS: Exploring
%   Realistic Neural Models with the GEneral NEural SImulation System.}  New York:
% TELOS/Springer--Verlag.

% [3] Hasselmo, M.E., Schnell, E.\ \& Barkai, E.\ (1995) Dynamics of learning and
% recall at excitatory recurrent synapses and cholinergic modulation in rat
% hippocampal region CA3. {\it Journal of Neuroscience} {\bf 15}(7):5249-5262.
% }

%%%%%%%%%%%%%%%%%%%%%%%%%%%%%%%%%%%%%%%%%%%%%%%%%%%%%%%%%%%%
\bibliographystyle{plain}
\bibliography{main}

\clearpage
\setcounter{page}{1}
% \maketitlesupplementary
\appendix
% \begin{figure*}
%     \centering
%     \includegraphics[width=\textwidth]{images/cvpr24_vis_vqa_v2.drawio.pdf}
%     \caption{Visualization of Coherence VQA on different models. The numbers after each questions are the probability of the model answering correctly about the corresponding question. \textcolor{red}{Red} numbers indicates the wrong answers. }
%     \label{fig:vis-coherence VQA}
% \end{figure*}
\section{Datasets Details}
\label{appdx: dataset details}
Table~\ref{tab:dialogben_stats} lists the detailed data composition of the benchmark.
\begin{table}[b!]
%\vspace{-2mm}
\begin{centering} 
\par\end{centering}
%\vspace{-4mm}
\begin{center}
% \begin{small}
\begin{sc}
\resizebox{\linewidth}{!}{
\begin{tabular}{cccc|cccc|cccc}

\toprule 
\multicolumn{4}{c}{Round1 (\%)}  & \multicolumn{4}{c}{Round2 (\%)} & \multicolumn{4}{c}{Round3 (\%)} \tabularnewline
\midrule
% T2T & T2I & I+T2T & I+T2I & T2T & T2I & I+T2T & I+T2I & T2T & T2I & I+T2T & I+T2I \tabularnewline
\textit{T}$\to$ \textit{T} & \textit{T}$\to$ \textit{I} & \textit{I+T}$\to$ \textit{T} & \textit{I+T}$\to$ \textit{I} & \textit{T}$\to$ \textit{T} & \textit{T}$\to$ \textit{I} & \textit{I+T}$\to$ \textit{T} & \textit{I+T}$\to$ \textit{I} & \textit{T}$\to$ \textit{T} & \textit{T}$\to$ \textit{I} & \textit{I+T}$\to$ \textit{T} & \textit{I+T}$\to$ \textit{I} 
\tabularnewline
  \midrule
  24.6 & 28.0 & 23.7& 23.7 & 24.1 & 27.6 & 23.3 & 25.0 & 24.2& 23.9& 24.5 & 27.2  \tabularnewline

\bottomrule
\end{tabular}
}
\vspace*{1mm}
\caption{\label{tab:data distribution}{The data proportion for each modality switching scenario in each round. The input modalities lie before $\to$ and output modalities lie after, where \textit{I},\textit{T} represent image and text respectively. }}
\label{tab:dialogben_stats}
\end{sc}
% \end{small}
\end{center}
% \vspace{-2em}
\end{table}

% \hmb{draw the pipeline of DialogGen including the data collection}

As stated in Sec.\ref{sec:exp}, our training datatset consists of $D_o$, $D_p$, $D_{pm}$ and DialogGen-train. In Tab.~\ref{tab: train_data_stat} we summarize the statistics of the components of the whole training datasets and in Example~\ref{exmp:IT_data} and Example~\ref{exmp:IT_data_dia} we give some examples of the instruction tuning data we used during our training.

\begin{exmp}{Instruction Tuning Data}{IT_data}
    \small
    % History:\\
    \bm{$D_o$}: \\ 
    \textbf{user}:\includegraphics[width=0.5\textwidth]{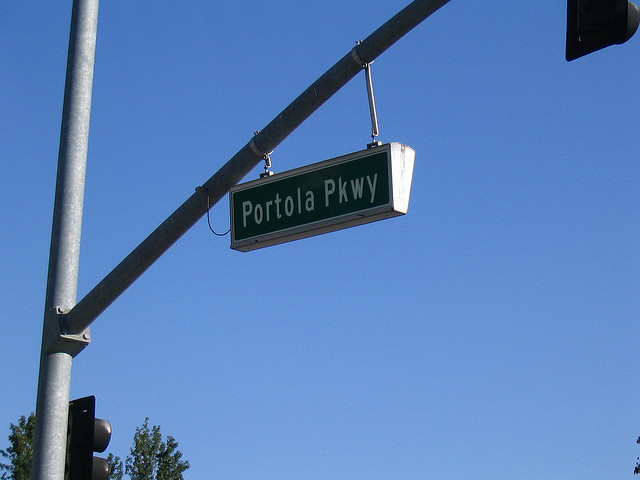}What is hanging from the traffic light pole?\\
    \textbf{assistant}: A green street sign is hanging from the traffic light pole.\\
    \\
    \bm{$D_p$}: \\
    \textbf{user}: I want to see a squirrel in the forest. \\
    \textbf{assistant}: \textless{}draw\textgreater{}A lively squirrel jumping in the forest, the background is dense forest, the lens is close-up, the style is cartoon style.\\
    \\
    \bm{$D_{pm}$}: \\
    \textbf{user}:\includegraphics[width=0.5\textwidth]{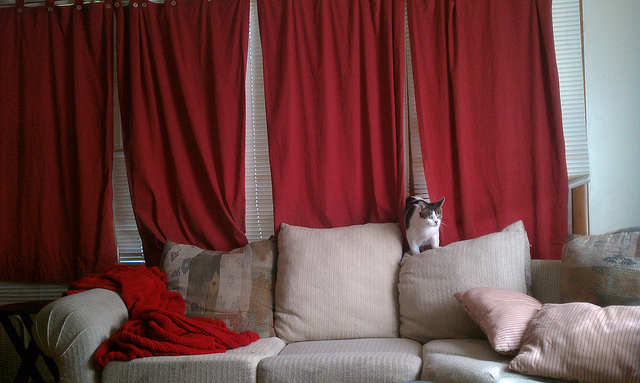}What is the cat doing in the image?\\
    \textbf{assistant}: The cat is coming out from some curtains onto the couch and is sitting or standing on top of it.\\
    \textbf{user}: Design a logo that reflects environmentalism.\\
    \textbf{assistant}: \textless{}draw\textgreater{}A circular logo, the main body is a hand in the care of a young sapling, the background is light green symbol of environmental protection and peace, viewing angle for squint, showing the harmony between human and nature, the style is simple style.\\
    \\
    % \textbf{DialogGen-train}:\\
    % \textbf{user}: I love fast trains. Can you draw me a picture?\\
    % \textbf{assistant}: \textless{}draw\textgreater{}A black steam train gallops along the tracks. The background is a forest in fog. The lens is close-up. The style is realistic.\\
    % \textbf{user}: Draw the train in cartoon style.\\
    % \textbf{assistant}: \textless{}draw\textgreater{}A black steam train gallops along the railway tracks. The background is a forest in fog. The camera is close-up. The style is cartoon style.\\
    % \textbf{user}: I prefer smooth lines, so try the line drawing style. \\
    % \textbf{assistant}: \textless{}draw\textgreater{}A black steam train gallops along the tracks. The background is a forest in fog. The lens is close-up. The style is line drawing. \\

\end{exmp}
\begin{exmp}{Instruction Tuning Data}{IT_data_dia}
    \small
    % History:\\
    
    \textbf{DialogGen-train}:\\
    Sample 1:\\
    \textbf{user}: I love fast trains. Can you draw me a picture?\\
    \textbf{assistant}: \textless{}draw\textgreater{}A black steam train gallops along the tracks. The background is a forest in fog. The lens is close-up. The style is realistic.\\
    \textbf{user}: Draw the train in cartoon style.\\
    \textbf{assistant}: \textless{}draw\textgreater{}A black steam train gallops along the railway tracks. The background is a forest in fog. The camera is close-up. The style is cartoon style.\\
    \textbf{user}: I prefer smooth lines, so try the line drawing style. \\
    \textbf{assistant}: \textless{}draw\textgreater{}A black steam train gallops along the tracks. The background is a forest in fog. The lens is close-up. The style is line drawing. \\
    \\
    Sample 2:\\
    \textbf{user}: \includegraphics[width=0.5\textwidth]{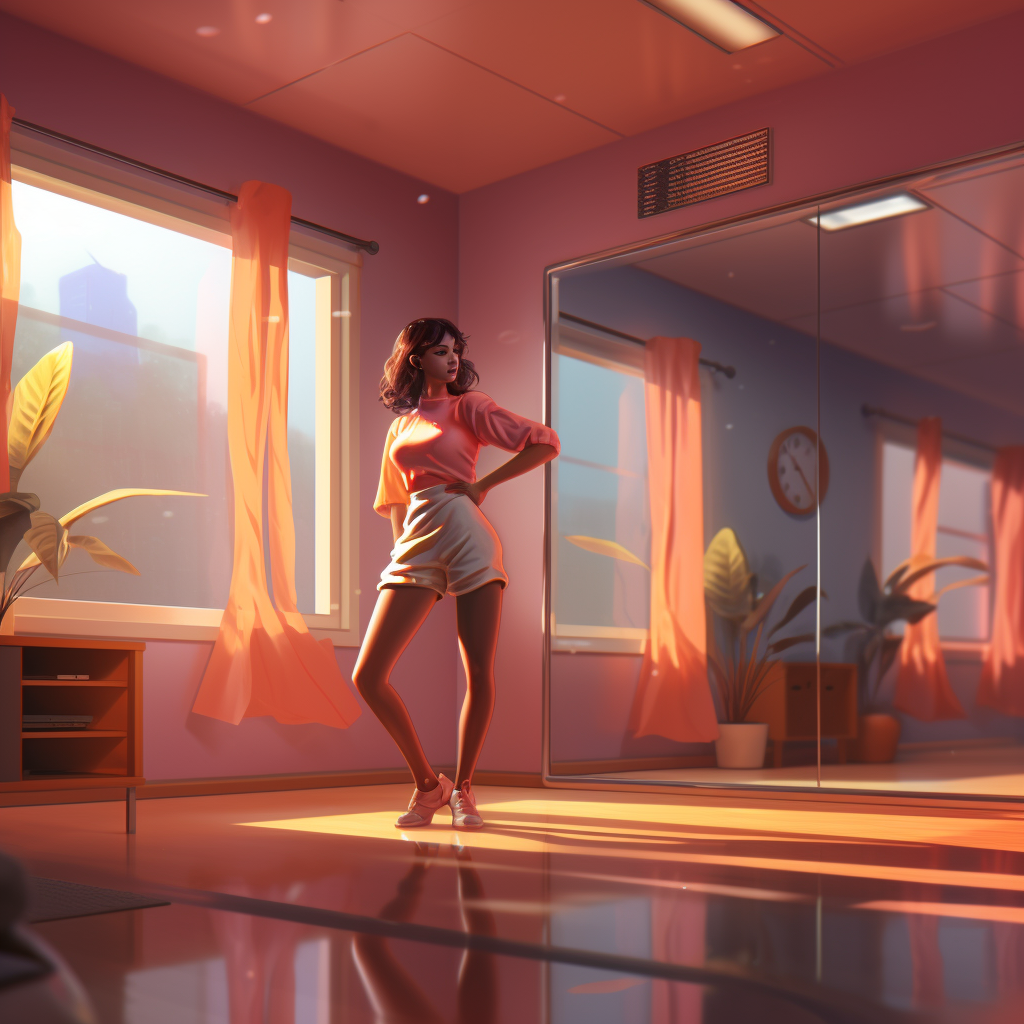}What is the woman in the picture doing?\\
    \textbf{assistant}: The woman in the picture is practicing dance.\\
    \textbf{user}: Can you add a little girl watching a girl dance to this picture and add some interactive elements?\\
    \textbf{assistant}: \textless{}draw\textgreater{}A woman dressed in a dance outfit was practicing dance in the room. Beside her sat a little girl with wide eyes, as if she was watching intently. The background was a simple and bright dance practice room. The camera was in the middle ground, and the style was realistic.\\
    \textbf{user}: How does the little girl feel, how does she feel about the dance?\\
    \textbf{assistant}: The little girl loved this dance and envied the dancing women. She seemed to be imagining that she could dance so beautifully in the future.

\end{exmp}

\section{Coherence VQA}
\label{appdx: vqa}
We provide visualization of Coherence VQA Score in Fig.~\ref{fig:vqa_result}. Here again DialogGen-X is DialogGen equipped with our closed source T2I model, and DialogGen-SD is the one equipped with  Stable Diffusion v1.5. NExt-GPT~\cite{wu2023next} and SEED-LLaMA~\cite{ge2023making} are two comparative models in the setting of Multi-modal Interactive Dialogue System.

As can be seen in the figure, in the first turn of the dialogue, both DialogGen-X and DialogGen-SD generate correct images except for a flaw in the action \textbf{jumping} for DialogGen-SD. While both NExT-GPT and SEED-LLaMA make two mistakes and the coherence VQA score correctly capture the mistakes. In the second turn of the dialogue, DialogGen-SD misses the action \textbf{jumping} and both NExT-GPT and SEED-LLaMA make two mistakes. Looking more closely we can find the DialogGen-X have a relatively better consistent main body.
Therefore, we consider coherence VQA score to be a suitable proxy that reflect to what extend the output image satisfy the user's need.

% Moreover, we also provide human evaluation on a small subset of $k=30$ examples whether all the models in Tab~\ref{tab: coherence_vqa} successfully identifies the correct intention of the user's as drawing and generating the corresponding images. For each of examples $i$ with corresponding $n_i$ questions, we ask the human annotators to give scores $\{s_{ij}\}_{j=1}^{n_i}$ to measure how likely the answer to these questions are true, from $s_{ij}=5$ being the most likely to $s_{ij}=1$ being the least likely. Afterwards, we take the average these scores to be the final human evaluation score $s_h=\frac{\sum\limits_{ij}s_{ij}}{\sum\limits_{i}n_i}$. And finally we scale it into the interval between 0 and 1 to have a more intuitive comparison between coherence VQA.

\begin{table*}[t!]

\small
    \setlength{\tabcolsep}{12pt}

    \centering
    \resizebox{.99\linewidth}{!}{
    \begin{tabular}{l|p{12cm} c}
    \toprule
           Data  & Descriptions & Size    \\
           \midrule
         \multirow{2}{*}{$D_o$} & It contains open-sourced uni-modal single-turn conversations like alpaca~\cite{alpaca} and multi-modal multi-turn conversations like LLaVA~\cite{chen2023llava}  &  \multirow{2}{*}{8.8K}\\
         \midrule
           % SEED-LLaMA & xx \\
           $D_p$ & It contains carefully generated single-turn text to drawing prompts conversations. & 3.6K \\
           \midrule
         \multirow{2}{*}{$D_{pm}$}  &It contains randomly concatenated samples from $D_o$ and $D_p$ to form multi-turn conversations that doesn't necessarily have the coherent semantics. & \multirow{2}{*}{5K}  \\
         \midrule
        \multirow{2}{*}{DialogBen-train}  & It contains carefully generated three-turn semantically coherent conversations. The input involve texts and images and the out involves texts and images (transformed drawing prompts) & \multirow{2}{*}{4.5K}  \\
    \bottomrule
    \end{tabular}
    }
    % \vspace{-3mm}
    \caption{Statistics of the instruction tuning datasets of DialogGen }
    % \vspace{-5mm}
    \label{tab: train_data_stat}
\end{table*}

\section{More visualization results}
\begin{figure}
    \centering
    \includegraphics[width=\textwidth]{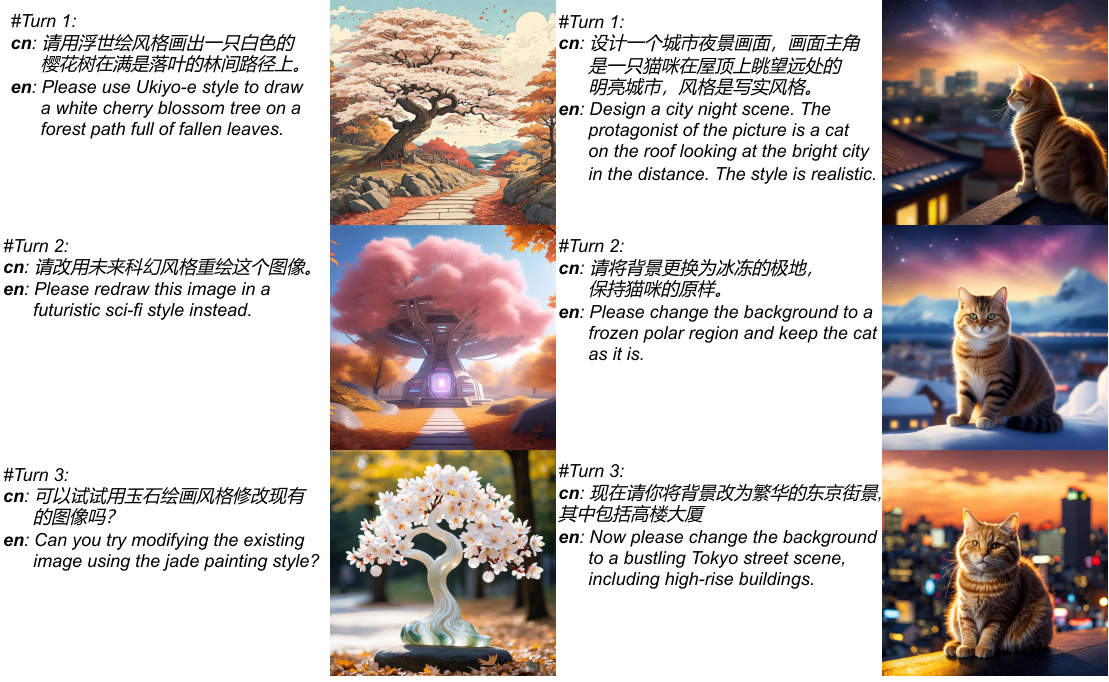}
    \caption{Visualization results of inference result of DialogGen-X}
    \label{fig:appdx vis 1}
\end{figure}

\begin{figure}
    \centering
    \includegraphics[width=\textwidth]{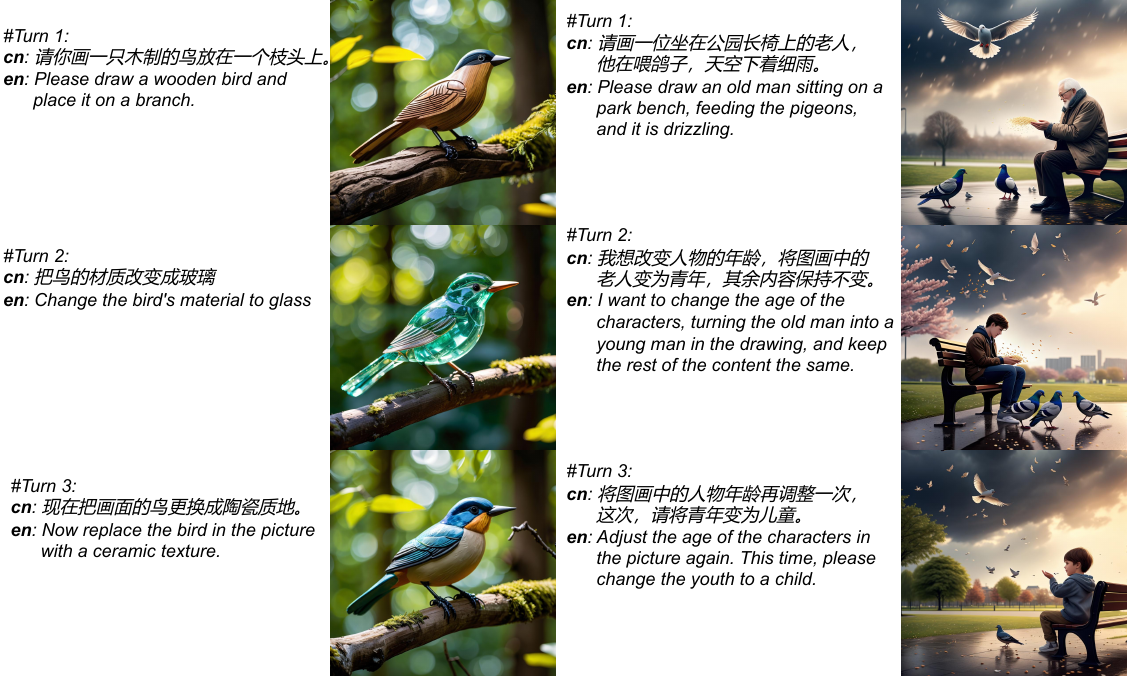}
    \caption{Visualization results of inference result of DialogGen-X}
    \label{fig:appdx vis 2}
\end{figure}

We provide more visualization results in Fig.~\ref{fig:appdx vis 1} and Fig.~\ref{fig:appdx vis 2}.

\begin{exmp}{Prompt For Generating Corrections}{correction_prompt}
\label{prompt for error correction}
    \small

    You are an expert in providing feedback on the output of an Multi-modal Large Language Model that supports multi-turn multi-modal dialog and image generation, the behaviour of the assitant is listed as follows:
    
1. The input may contains images. It is specailly identified by \textless{}img\textgreater{} and \textless{}/img\textgreater{}  where the detailed description of the image lies between them.

2. For the input of an user, it will first identify the whether the intention of the user is draw or talk, if it is to draw it will append \textless{}draw\textgreater{} before the output.

3. If the user intention is to talk, then the output of the assistant in current turn should be coherent in response to the user query.

4. If the user intention is to draw, then the output of the assistant in current turn should be \textless{}draw\textgreater{} and a detailed description prompt in response to the user's need. The prompt should be able to be directly sent into the image generation model to produce high quality image.

You are given the history conversation H, user query in current turn Q, output of the assistant in current turn $R_{pre}$ and a suitable output given by human expert $R_{gt}$. 

You need to make the judgements as following rules:

1. According to the history and the current user query, make judgements on whether the model correctly identifies the use's intention.

2. If the intention of the user is to talk, make judgements on whether the output of the assistant following the instructions of the user coherently.

3. If the intention of the user is to draw, make judgements on whether the output prompt is indeed a descriptive text that is suitable for a image generation model to generate images.

If the output satisfy all the three criteria, output '\#\#\#Correct'. If not, output '\#\#\#Wrong\#\#\#' and the specific criteria it violates followed by an explanation and provide a correct output for current user query..\\
    
    % \textbf{Question:} \\
    % \textbf{Original Output:}\\
    % xxxxxxxxx\\
    % \textbf{Correction:} ......\\
    % \textbf{Correct Solution:}\\
    % ...\\

    (3 in-context examples)\\

    \textbf{Question:} $q_{i}$\\
    \textbf{Original Output:} $r_{i}$\\
    \textbf{Correction:} \{Completion\}

\end{exmp}
 
% \section{Qualitative Examples}

\end{document}